\documentclass{article}

\usepackage{microtype}
\usepackage{graphicx}
\usepackage{subfigure}
\usepackage{booktabs} 

\usepackage{hyperref}
\usepackage{cite}

\usepackage[accepted]{icml2018}
\newcommand*{\affaddr}[1]{#1} 
\newcommand*{\email}[1]{\texttt{#1}}

\begin{document}

\date{}
\title{TUNet: Incorporating segmentation maps to improve classification}

\author{%
Yijun Tian\\
\affaddr{Department of Computer Science}\\
\affaddr{New York University}\\
\email{eddie.tian@nyu.edu}\\
}

\maketitle

\icmlkeywords{Deep Learning, Classification, Segmentation}

\vskip 0.3in

\begin{abstract}
Determining the localization of specific protein in human cells is important for understanding cellular functions and biological processes of underlying diseases. Among imaging techniques, high-throughput fluorescence microscopy imaging is an efficient biotechnology to stain the protein of interest in a cell. In this work, we present a novel classification model Twin U-Net (TUNet) for processing and classifying the belonging of protein in the Atlas images. Several notable Deep Learning models including GoogleNet and Resnet have been employed for comparison. Results have shown that our system obtaining competitive performance.

\end{abstract}

\section{Introduction}

Spatial partitioning of biological functions is a phenomenon fundamental to life and Proteins are “the doers” in the human cell, executing many functions that together enable life. Moreover, the localization of protein is strongly associated with cellular dysfunction and disease. Thus, knowledge of the spatial distribution of proteins at a subcellular level is essential for understanding protein function, interactions, and cellular mechanisms (Thul et al., 2017). 

Images visualizing proteins in cells are commonly used for biomedical research, and these cells could hold the key for the next breakthrough in medicine. A reliable biotechnology to localize proteins is high-throughput fluorescence microscopy imaging (HTI), providing high-quality protein analysis image (Pepperkok and Ellenberg, 2006). With which we can localize proteins straightforward (Swamidoss et al., 2013). Therefore, the need is greater than ever for automating biomedical image analysis to accelerate the understanding of human cells and disease.

\begin{figure}[ht]
\vskip 0.1in
\begin{center}
\centerline{\includegraphics[width=0.9\columnwidth]{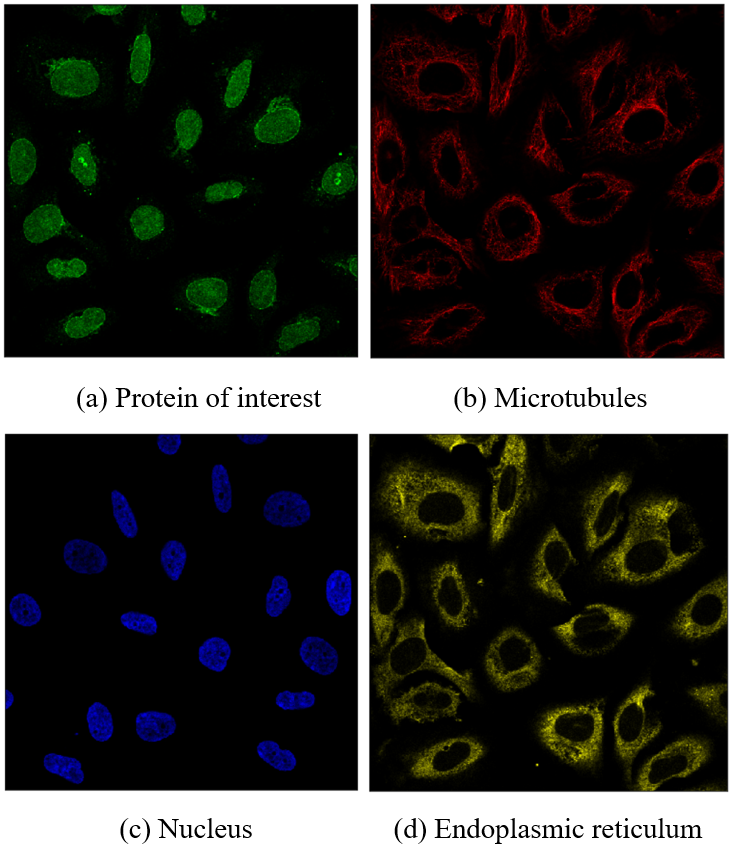}}
\caption{Exemplary samples from the dataset.}
\end{center}
\vskip -0.4in
\end{figure}

Historically, classification of proteins has been limited to single patterns in one or a few cell types. However, in order to fully understand the complexity of the human cell, models must classify mixed patterns across a range of different human cells. Therefore, with Deep neural networks (LeCun et al. 2015; Schmidhuber 2015) becoming popular for image analysis tasks, scholars are trying to utilize it in biological domains (Tan et al. 2015; Angermueller et al. 2016) and predict to which subcellular compartment a protein belongs (Sønderby et al.,2015; Pärnamaa 2017; Armenteros 2017). Among all of the networks, the presentation of Convolutional Neural Networks (CNNs) plays a dominant role (Krizhevsky et al., 2012) and many notable models have been proposed in the field of image classification, e.g., GoogleNet (Szegedy, C. 2014), ResNet (He et al., 2015) and DenseNet (Huang et al., 2017). Recently, human experts successfully apply CNNs in the field of biological and medical images, making great breakthrough in the tasks of skin cancer detection (Esteva et al., 2017) and lesions in mammograms (D. Ribli, 2018). Here, we aim at a more challenging task in which proteins have to be classified to 28 classes with multiple possible labels per image. 
\begin{figure*}[ht]
\vskip 0.1in
\begin{center}
\centerline{\includegraphics[width=\linewidth,height=5.8cm]{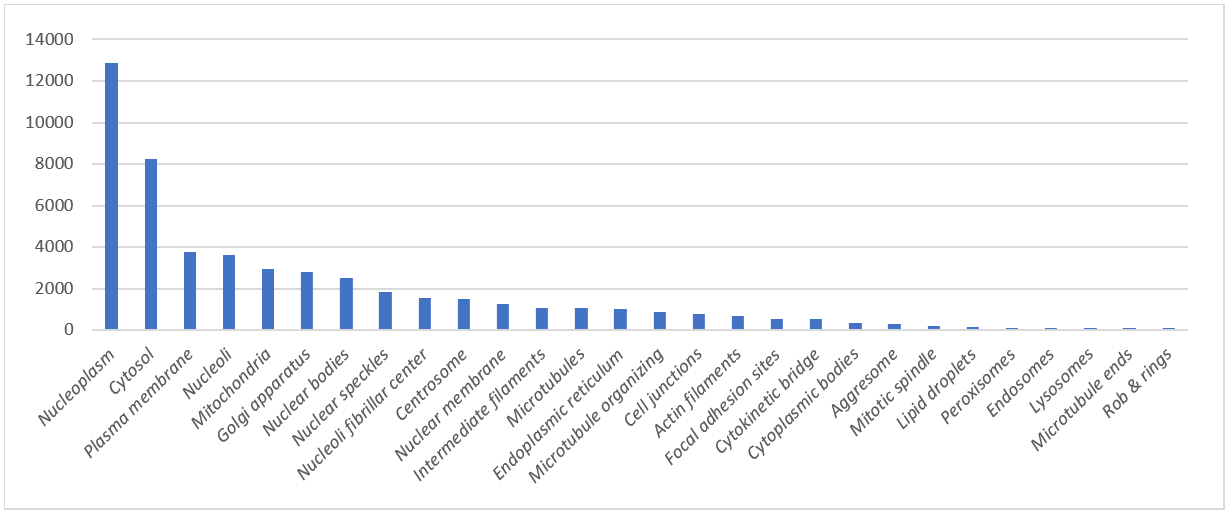}}
\caption{Frequency of protein location.}
\end{center}
\vskip -0.2in
\end{figure*}

\begin{table*}[ht]
\vskip 0.1in
\begin{center}
\centerline{\includegraphics[width=\linewidth]{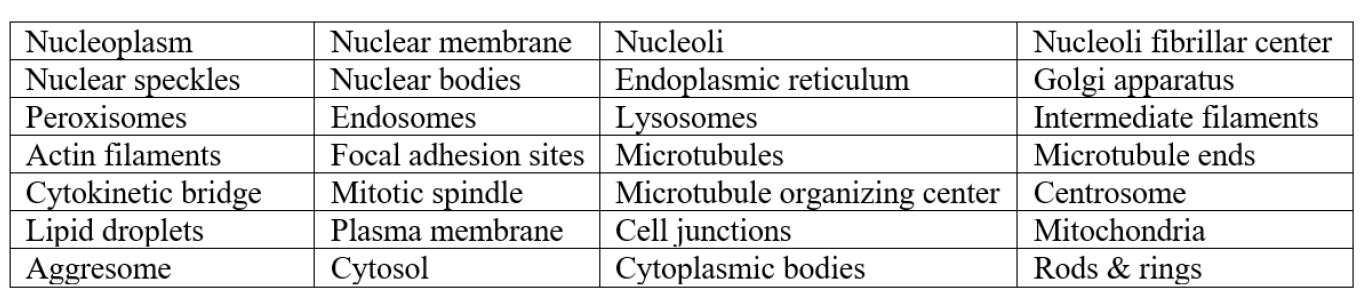}}
\caption{Locations of human cell protein.}
\end{center}
\vskip -0.3in
\end{table*}

Protein localization based on biomedical microscopy images incurs a special problem in machine learning for healthcare, namely, how to achieve great performance with not sufficient training data. In the field of healthcare, Medical imaging usually has its own feature: insufficient data. Thus, how to make full use of limited data to extract information becomes a problem. Therefore, we propose a method to help the model focus more on the target by incorporating the segmentation information. In this way, the segmentation mask generated by the data itself can also improve the classification effect. Thus, adaptions of our architecture is a good way to deal with the insufficient data problem.

In this work, we propose a framework to classify mixed patterns of proteins in microscope images. We assess and compare its performance on the largest available public dataset of high resolution HTI data in the field of subcellular protein localization in human cells. By comparing several different Deep Learning model’s performance, we show TUNet achieve state-of-art performance on this task.

\section{Dataset}

The model we proposed and the related experiments are all based on the dataset released by the International Society for Advancement of Cytometry (ISAC). It contains 31,072 samples taken from the Cell Atlas (Thul et al., 2017) which is part of the Human Protein Atlas. These images Maps the in situ localization of human proteins at a single-cell level to 28 subcellular structures. Therefore, the Cell Atlas is a rich source of information for protein classification.

In the dataset, each sample image contains two fold we can use. One is the full size original images and the other is scaled images. The former one is made up of TIFF files, mixture of 2048x2048 and 3072x3072 resolution, while the latter one contains only 512x512 PNG files.

For each sample, the label indicates the localization of protein organelle. There are in total 28 different labels present in the dataset, acquired in a highly standardized way using one imaging modality (confocal microscopy). However, the dataset comprises 28 different cell types of highly different morphology, which affect the protein patterns of the different organelles. Cell types are listed in Table 1. All image samples are represented by four filters(stored as individual files), the protein of interest (green) plus three cellular landmarks: nucleus (blue), microtubules (red), endoplasmic reticulum (yellow). Hence, in the landmarks, we use green filter to predict the label, and use others as references. We show image example in Figure 1.

\begin{figure}[ht]
\vskip 0.1in
\begin{center}
\centerline{\includegraphics[width=\columnwidth]{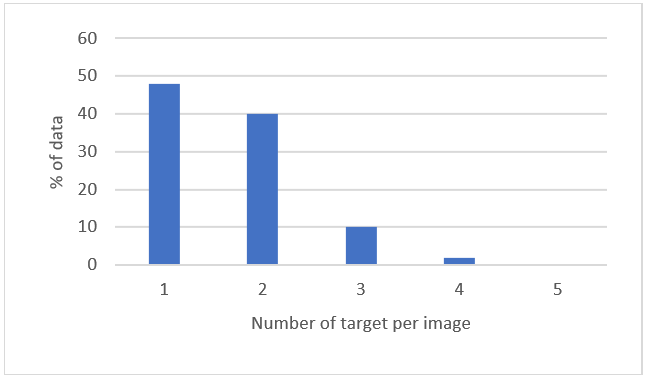}}
\caption{Statistics of labels per image.}
\end{center}
\vskip -0.4in
\end{figure}

\begin{figure}[ht]
\vskip 0.1in
\begin{center}
\centerline{\includegraphics[width=\columnwidth]{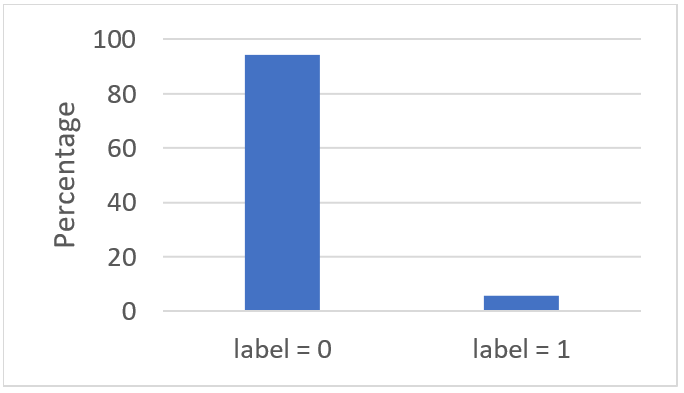}}
\caption{Count of true target label.}
\end{center}
\vskip -0.4in
\end{figure}

\subsection{Data Analyzing}
Before we build our model, we need to analyze the dataset. The first question is the distribution. Since the uniform distribution or normal distribution makes a great impact to our model design. We visualize the most occurred protein’s label in Figure 2. We can see that most common protein structures belong to coarse-grained cellular components like the plasma membrane, the cytosol and the nucleus. 

In contrast small components like the lipid droplets, peroxisomes, endosomes, lysosomes, microtubule ends, rods and rings are very seldom in our train data. For these classes the prediction will be very difficult as we have only a few examples that may not cover all variabilities and as our model probably will be confused during learning process by the major classes. Due to this confusion, we will make less accurate predictions on the minor classes. Consequently, accuracy is not the right score here to measure our model’s performance. 

In addition, given the 28 different labels, we want to know how many labels occurs most in one image. We count the number of target appeared in each image. Figure 3 shows the distribution of target number per image. We can conclude that most train images only have one or two target labels and more than three targets are very seldom.

The class each protein image correspondent is also very sparse. We plot the distribution in Figure 4. We can see that most of our targets are filled with zero, because of the imbalanced distribution of data labels between zero and one. This indicates to the absence of corresponding target proteins. 

For the correlation between our data, we want to know whether the position of protein occurs independently or relatedly. We use heat map to show the relationship between labels in Appendix. We can see that many targets only have very slight correlations. However, endosomes and lysosomes often occur together and sometimes seem to be located at the endoplasmic reticulum.

In addition, we find that the mitotic spindle often comes together with the Cytokinetic bridge. This makes sense, as both are participants for cellular division. In addition, in this process microtubules and their ends are active and participate as well. Consequently, we find a positive correlation between these targets. Further study about the seldom targets shows that it also have some kind of grouping with other targets. It reveals where the protein structure seems to be located. For example, we can see that rods and rings have some relationship with the nucleus whereas peroxisomes may be located in the nucleus as well as in the cytosol.

\begin{figure*}[ht]
\vskip 0.1in
\begin{center}
\centerline{\includegraphics[width=0.8\linewidth]{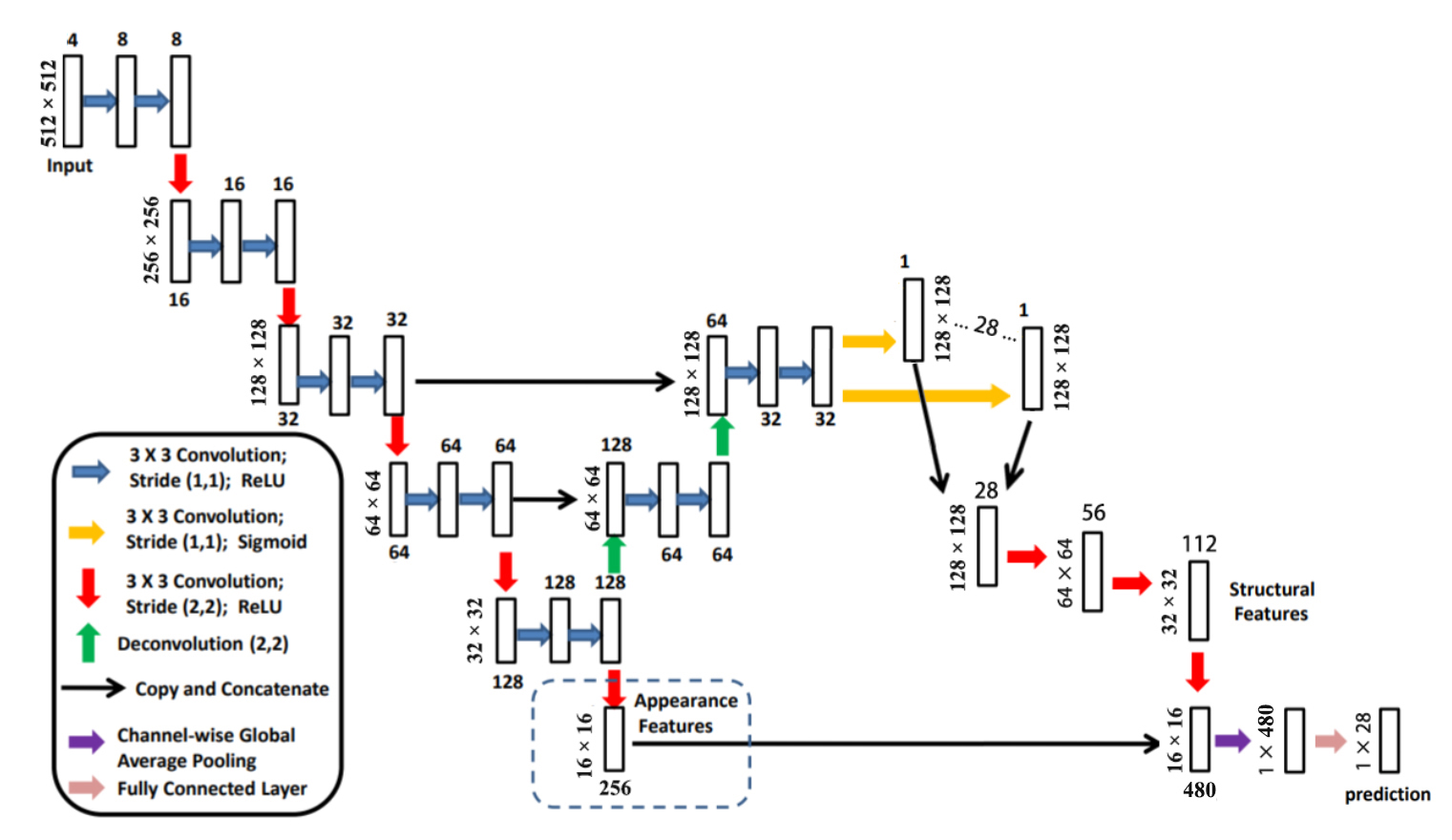}}
\caption{TUNet.}
\end{center}
\vskip -0.2in
\end{figure*}

\section{Model}

An outline of the proposed method is depicted in Appendix. Given an input four channel image, a ground truth mask is generated by threshold green channel. If the number of channel corresponds to the number of class label, then we set the mask with its’ green channel mask, depicting the protein of interest. Otherwise, we set mask to totally black, indicating there are no area the model need to focus on. Then, the image is provided as an input to the Multi-task Convolutional Neural Network architecture, TUNet, while the mask we generated before as label of the segmentation part of model. The model train the mask generation process and the predicting part jointly, providing an image level class prediction at final. Further details of TUNet is discussed in Section 3.1. The binary segmentation masks obtained from TUNet are further refined in a post-processing step, including remove noise to further improve their accuracy.

\subsection{TUNet}

Figure 5 depicts the architecture of the proposed Multi-task classification CNN network, TUNet. We resized the input image to 512 dimension and fed it into a convolutional network similar to the U-net (Ronneberger 2015). It consists of an encoder path (left side), a decoder path (middle) and a contracting path (right side). Additional skip connections are given between the corresponding encoder and decoder layers. The final output of the decoder network is fed into separate convolution layers with a sigmoid activation function to obtain the output segmentation masks for different class. In the classification part, the model combines the appearance features (corresponding to image information) and the structural features (corresponding to the segmentation mask of image). The image appearance features are extracted by reusing the output of encoder path and apply it to one more convolutional layer with stride (2, 2) and ReLU activation function. With the aim of increasing prediction performance, we also incorporate structural features, which is obtained by a series of convolutional layers with stride (1, 1) and (2, 2). In the end, the appearance features and the structural features are concatenated to produce an informative feature vector. The model will output the final prediction score by applying a channel-wise global average pooling and a fully connected layer.

\begin{table*}[ht]
\vskip 0.1in
\begin{center}
\centerline{\includegraphics[width=\linewidth]{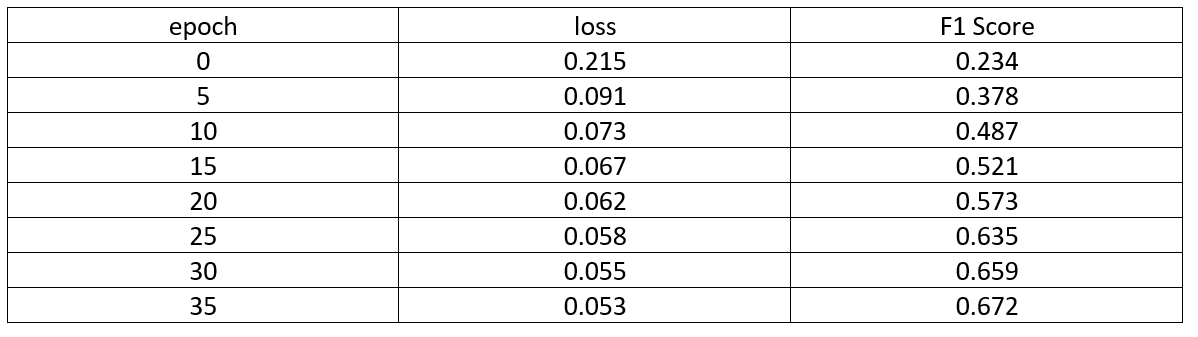}}
\caption{Performance on validation set.}
\end{center}
\vskip -0.2in
\end{table*}

\begin{table*}[ht]
\vskip 0.1in
\begin{center}
\centerline{\includegraphics[width=\linewidth]{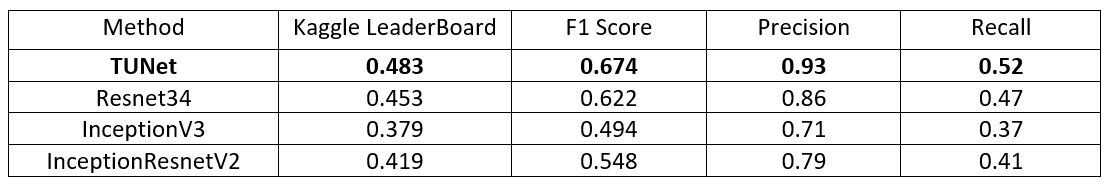}}
\caption{Performance comparison of different CNN architectures.}
\end{center}
\vskip -0.2in
\end{table*}

\subsection{Implementation details}

A loss function has to be defined for the multi-task CNN network. TUNet use focal loss for the image level classification task and dice coefficient for the measure of the generated segmentation masks performance. Given the purpose of focusing more on classifying task more than segmenting task, we set different weight to the subtask loss and combining them together as

\begin{equation}
Loss = \alpha\times L_{seg}+(1-\alpha)\times L_{cls}
\end{equation}

where we set $\alpha$ equals to 0.4 in our experiment, letting the model pay more attention to classification part. However, one of the challenges of designing the loss function is strong data imbalance. Some classes, like “Nucleoplasm”, are very common, while there is a number of rare classes, like “Endosomes” are very common, while there is a number of rare classes, like “Endosomes”, “Lysosomes”, and “Rods \& rings”. Therefore, it is crucial to use a loss function that accounts for it. Recently proposed focal loss (D. Ribli, 2018) has revolutionized object localization method in 2017. It was designed to address the issue of strong data imbalance, demonstrating amazing results on datasets with imbalance level 1:10-1000. Focal loss puts more attention to incorrectly classified examples by reducing the relative contribution of well-classified ones. In this case, the model is really trying to correct errors rather than decrease the total loss by a slight improvement of already correctly classified examples. Therefore, we use focal loss defined as

\begin{equation}
L_{cls} = -(1-p_t)^{\gamma}log(p_t)
\end{equation}

Here $\gamma$ are focusing parameter, controlling the
strength of the modulating term and $p$ is the estimated probability for each class. For the segmentation part, we use dice loss to evaluate our model, which is defined as 
\begin{equation}
L_{seg} = 1-Dice(R, Y)
\end{equation}
where $R$ denotes the segmentation masks generated by the network and $Y$ represents the ground truth segmentation masks.The Dice metric was defined similar to [5] as
\begin{equation}
Dice(R, Y) = \frac{2\sum_{i}r_i\times y_i}{\sum_{i}r_i+\sum_{i}y_i}
\end{equation}
where $r_i$ and $y_i$ represents the value at the $i^{th}$ pixel in $R$ and $Y$ respectively and the summations run over all the pixels in the image.

For the evaluation metrics, we can see that the label’s space is very sparse from Figure 4. Thus, we may not able to use accuracy as our validation metric. In order to solve the problem of unbalanced distribution, we introduced $F1$ score. Since the Cell Atlas data are strongly imbalanced classified. We prefer to use $F1$-macro score rather than $F1$-micro score, giving more weight to those categories with fewer samples.

\section{Experiment}

We use Adam as optimization function to update the parameters of TUNet automatically. The batch size was set to 64 and early stopping was used to terminate the training when the validation loss did not decrease. Since data augmentation plays a crucial role in the performance of model. We apply rotation, dihedral and lighting change to the initial image data. Since the segmentation masks generated by TUNet is binary pixel-labeling problems, with the aim of improving performance, we may use OpenCV library to remove the noise of mask. Moreover, all of the segmentation softmaps are binarized by setting threshold at 0.5. However, for the classification part, experiment shows that the threshold of 0.5 for final output is not a good choice. In this case, we ignore the property of data itself and the difference between classes. In order to help model output more accurate class choice, we let the threshold chosen automatically according to the characteristic of data instead of manually selected, namely, select each value based on the least square error between the class data and its label. In this way, every class has its own threshold for class decision. 

From 31057 training images, we randomly select 10\%, namely, 3105 images for validation set and use the rest of images for training set. We compared TUNet to other models like InceptionV3, InceptionResnetV2 and resnet34. Table 3 shows the comparison result. Experiment shows our model, TUNet achieves comparing performance with Resnet34 and far beyond the performance of InceptionV3 and InceptionResnetV2. With a higher precision score, TUNet are able to achieve better classification effect in the task of imbalanced data distribution. 

For the choice of learning rate, we begin with finding the optimal learning rate according to the loss. Increase of the loss indicates onset of divergence of training. The optimal learning rate lies in the vicinity of the minimum of the curve but before the onset of divergence. Here, we set the initial learning rate at 0.02. After that, we use learning rate annealing to train the model more precisely. Periodic learning rate increase followed by slow decrease drives the system out of steep minima towards broader ones that enhances the ability of the model to generalize and reduces overfitting. 

Table2 shows the relationship between training epoch, model loss and F1 score. The model tends to converge from 30 to 35 epoch. The loss start decreasing fast in the beginning, then after the model learning the characteristic of cell image, F1 score start increasing rapidly, showing strong learning ability of TUNet. We apply early stopping to terminate the training. To avoid overfitting, dropout is used at the end of contracting path and the final fully connected layer. 

\section{Conclusion}

In this work, we have explored a novel Multi-classification CNN architecture, TUNet, to overcome the challenges introduced by the nature of HTI data such as imbalanced data and weak labels. By fully utilizing the image information, TUNet can jointly segmenting different class and predicting the class belonging. With the aim of reducing the computational requirements and improving the generalizability of the learned features, the features of the CNN are shared across the segmentation task and classification task. In a large study comparing convolutional neural architectures for human protein localization, through incorporating the segmentation mask of cell image, which contains abundant information of cellular location, TUNet shows competitive performance, while remains less parameters and fewer training time.

\nocite{Thuleaal3321}
\nocite{Uhlen2010}
\nocite{Pepperkok2006}
\nocite{Swamidoss2013}
\nocite{NIPS2012_4824}
\nocite{2014arXiv1409.4842S}
\nocite{2015arXiv151200567S}
\nocite{2015arXiv151203385H}
\nocite{2016arXiv160806993H}
\nocite{Esteva2017}
\nocite{Ribli2018}
\nocite{2017arXiv170802002L}
\nocite{10.1007/978-3-319-21233-3_6}
\nocite{LeCun2015}
\nocite{Schmidhuber2015DeepLI}
\nocite{Tan2015}
\nocite{Angermueller878}
\nocite{Parnamaa1385}
\nocite{doi:10.1093/bioinformatics/btx431}
\nocite{2015arXiv150504597R}

\bibliography{example_paper}
\bibliographystyle{icml2018}

\end{document}